\documentclass{article}

\usepackage[final]{corl_2020} 

\usepackage{bm}
\usepackage{url}
\usepackage{ulem}
\usepackage{amsmath}
\usepackage{amssymb}
\usepackage{amsfonts}
\usepackage{booktabs}
\usepackage{multirow}
\usepackage{multicol}
\usepackage{graphicx}
\usepackage{algorithm}
\usepackage{algpseudocode}

\usepackage{subcaption}
\usepackage{stackengine}
\usepackage{scalerel}
\usepackage{graphicx}

\newcommand{\PegInsertion}{{\it Peg-Insertion\,}}

\newcommand{\PickPlace}{{\it Pick-and-Place\,\,}}

\newcommand{\Tref}[1]{Table~\ref{#1}}
\newcommand{\eref}[1]{Eq.~(\ref{#1})}

\newcommand{\fref}[1]{Fig.~\ref{#1}}
\newcommand{\Fref}[1]{Figure~\ref{#1}}
\newcommand{\sref}[1]{Sec.~\ref{#1}}

\title{Deep Reactive Planning in Dynamic Environments}
\author{%
  Kei Ota$^{1}$\thanks{Correspondence to:  \texttt{\href{mailto:Ota.Kei@ds.MistsubishiElectric.co.jp}{Ota.Kei@ds.MistsubishiElectric.co.jp}} and  \texttt{\href{mailto:jha@merl.com}{jha@merl.com}}} \qquad 
  Devesh K. Jha$^{2}$ \qquad 
  Tadashi Onishi$^{1}$ \qquad 
  Asako Kanezaki$^{3}$ \\
  \textbf{
  Yusuke Yoshiyasu $^{4}$ \qquad
  Yoko Sasaki$^{4}$ \qquad 
  Toshisada Mariyama$^{1}$ \qquad 
  Daniel Nikovski$^{2}$ \qquad
  } \\
  $^1$ Mitsubishi Electric ~~ $^2$MERL ~~ $^3$Tokyo Institute of Technology ~~ $^4$AIST
}

\begin{document}
    \maketitle

    \begin{figure}[h]
        \centering
        \includegraphics[width=\textwidth]{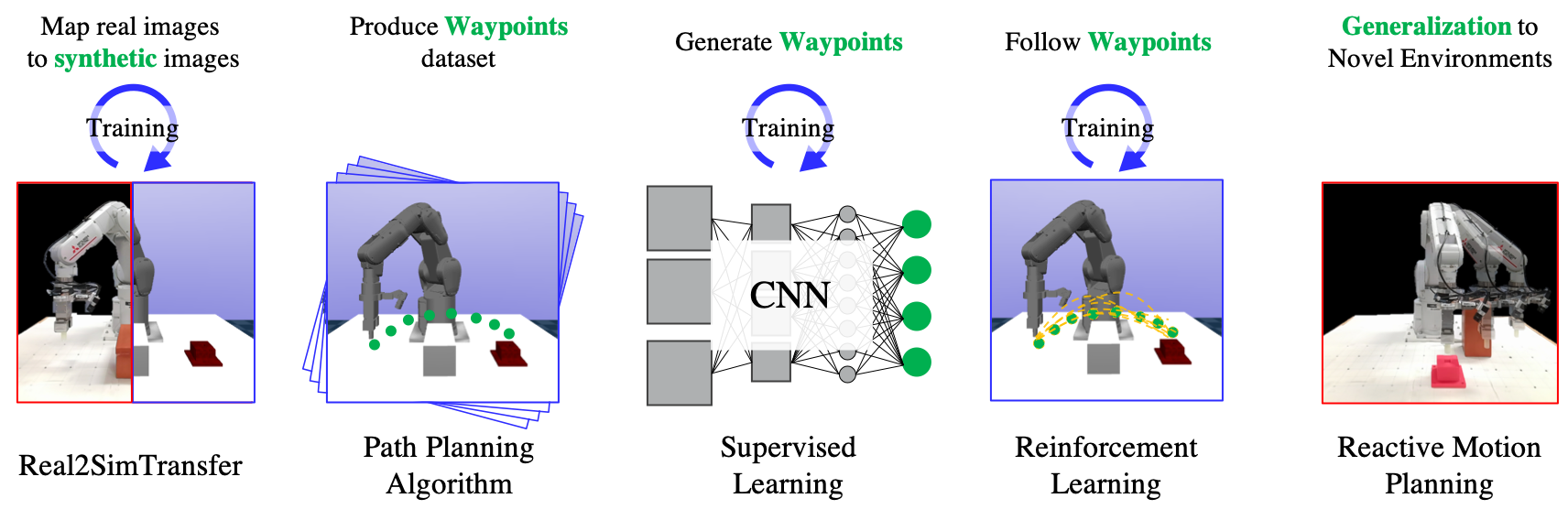}
        \caption{Our proposed agent learns an end-to-end reactive planning technique by combining traditional path planning algorithms, supervised learning (SL) and reinforcement learning (RL) algorithms in a synergistic way. A deep CNN is used to learn the sequence of waypoints obtained from a kinematic planning algorithm (e.g., a Bidirectional RRT*) given a depth image of the environment. The agent learns to follow arbitrary waypoints using path-conditioned RL, thus resulting in efficient exploration. We show that our trained agent can achieve good sample efficiency, as well as generalization to novel environments in simulation as well as real environments. The whole learning process is done in the simulator by learning a Real2Sim transfer function to make the training process efficient and suitable for robotic systems.}
        \label{fig:proposed_method}
    \end{figure}
    \begin{abstract}
The main novelty of the proposed approach is that it allows a robot to learn an end-to-end policy which can adapt to changes in the environment during execution. While goal conditioning of policies has been studied in the RL literature, such approaches are not easily extended to cases where the robot's goal can change during execution. This is something that humans are naturally able to do. However, it is difficult for robots to learn such \textit{reflexes} (i.e., to naturally respond to dynamic environments), especially when the goal location is not explicitly provided to the robot, and instead needs to be perceived through a vision sensor. In the current work, we present a method that can achieve such behavior by combining traditional kinematic planning, deep learning, and deep reinforcement learning in a synergistic fashion to generalize to arbitrary environments. We demonstrate the proposed approach for several reaching and pick-and-place tasks in simulation, as well as on a real system of a $6$-DoF industrial manipulator.
\end{abstract}


    \section{Introduction}\label{sec:introduction}

Deciding how to reach a goal state by executing a long sequence of actions in robotics and AI applications has traditionally been in the domain of automated planning, which is typically a slow, deliberative process that makes extensive use of knowledge about how the world works and how the agent can operate in it. Yet, humans and other living organisms demonstrate amazing ability to replan very quickly when the goal is moving, for example, when a predator is chasing prey or when a baseball player is trying to hit a fast-moving ball. This ability is likely to result not from just very fast execution of a basic planning procedure, but from skillful generalization over many previously computed and suitably cached solutions to instances of planning problems with variable goals. For a baseball player, these would be the many successful swings performed during target practice, over many variations of the trajectory of the incoming ball. While recently we have seen tremendous progress in reinforcement learning and deep learning, designing agents which can demonstrate such behavior still remains elusive. Our work is motivated by designing agents that can demonstrate such intelligent behavior by reacting to changes in an environment during execution.

We propose a learning-based approach to reactive planning, inspired by the way humans and animals perform it,  reacting to change in environments effortlessly due to their highly responsive reflexes. As long as the planning environment does not involve tight, complex obstacle-cluttered environment, most humans can swiftly change their plan and still perform a trajectory-centric task without having to replan the original trajectory. Within this approach, a central question is how generalization should take place, and what the output of the generalizing module should be. While it is tempting to output and generalize directly over motor commands, this might make the learning problem unnecessarily hard, requiring too many training samples. Instead, we propose to decouple the problem into a trajectory (re)-planning component, subject to generalization, and a trajectory following component, which remains constant and goal-independent (see \fref{fig:proposed_method}). 

The main contribution of the proposed work is to present a technique for training agents that can perform reactive tasks in dynamic environments in an end-to-end fashion. We synergistically bring together planning, supervised learning, and reinforcement learning to design controllers for robotic systems, which can use low-cost vision systems to perform trajectory-centric tasks. Furthermore, we use a real2sim transfer technique, which allows us to design policies only in simulation by learning a function that maps real images to simulated ones, and thus makes training more efficient. On an intuitive note, the proposed technique decouples the problem of trajectory planning and trajectory-centric control. More specifically, the trajectory planner is a neural network that takes as input an image of the environment and outputs a collision-free trajectory defined by a sequence of relatively few waypoints. This neural network is trained in a supervised learning fashion using kinematic trajectories generated in different environments using an off-line deliberative planner such as RRT. The trajectory-centric controller is a deep RL policy that takes as an input an arbitrary trajectory and allows the robot to follow it (see \fref{fig:proposed_method}). The training is done in a simulation environment by performing real2sim transfer for efficient training and the trained policy is transferred to the real system without further fine-tuning.
    \section{Related Work}\label{sec:related_work}
In this section, we review related work in open literature. 
Motion planning is one of the core and widely studied topics in robotics. The most widely used algorithms are sampling-based methods, such as rapidly exploring random trees (RRT) \cite{lavalle2006planning} and probabilistic roadmaps (PRM) \cite{kavralu1996probabilistic}. There are several other optimization-based techniques such as CHOMP and STOMP~\cite{Ratliff2009chomp, Kalakrishnan2011stomp}. While RRT-like approaches are not reactive per se, there are several approaches which use sampling-based algorithms for reactive planning~\cite{bruce2002real,gayle2007reactive}. However, all these traditional approaches require explicit detection and tracking of obstacles~\cite{kappler2018real-time, morrison2020learning}.

The combination of reference paths and RL has been widely researched \cite{faust2018prm,Thomas_2018,chiang2019learning,ota2019trajectory}.
In \cite{faust2018prm,chiang2019learning}, Probabilistic Roadmaps (PRM)~\cite{kavralu1996probabilistic} are used to find reference paths, and RL is used for point-to-point navigation as a local planner for PRM.
In \cite{Thomas_2018}, a model-based RL agent is learned for autonomous robotic assembly by exploiting the prior knowledge in the form of CAD data, using a Guided Policy Search (GPS) approach to learn a trajectory-tracking controller. However, the GPS algorithm still produces very local policies and cannot generalize to changing environments. Another relevant set of ideas could be found in~\cite{lotjens2019safe}, where a model-based approach is used to predict movement of obstacles in the environment of the robot and an MPC approach is used to replan. This is done in our proposed technique, too, in a more implicit manner. Another related and widely successful algorithm that learns a visuomotor policy from several example trajectories could be found in~\cite{levine2016end, levine2013guided}. However, it is, by its nature, goal-centric, and cannot be easily extended for reactive planning.

Goal parameterization in RL has been studied extensively to design policies which can generalize to different goals~\cite{nasiriany2019planning, chang2019learning}. One of the main differences that we would like to point out is that goal parameterization in RL cannot achieve reactive planning as it would require an auxiliary function that implicitly or explicitly updates the goal states from an observation. In our proposed work, this is achieved as the path encodes the goal and automatically tries to avoid obstacles in the environment, too. The closest work to ours is \cite{ota2019trajectory,ota2020efficient}. \cite{ota2019trajectory} learns an RL agent that optimizes trajectory for a 6-DoF manipulator arm. Our approach, however, can deal with changes in the environment by conditioning an RL agent with a reference path which is also generated based on an observation of the current environment.
\cite{ota2020efficient} considers a similar setting for 2-dimensional robot navigation tasks only in simulation.
Our method, however, considers higher-dimensional path planning, and is evaluated on real systems as well as simulated ones. Moreover, we try to minimize data collection in the real systems by additionally introducing the Real2Sim transfer pipeline, which is learned in a efficient way by using a common approach for domain adaptation~\cite{shrivastava2017learning,hoffman2018cycada,bousmalis2017unsupervised,rao2020rl}, and thus becomes more challenging than~\cite{ota2020efficient}.

    \section{Method}
    \begin{figure}[t]
        \centering
        \includegraphics[width=0.99\textwidth]{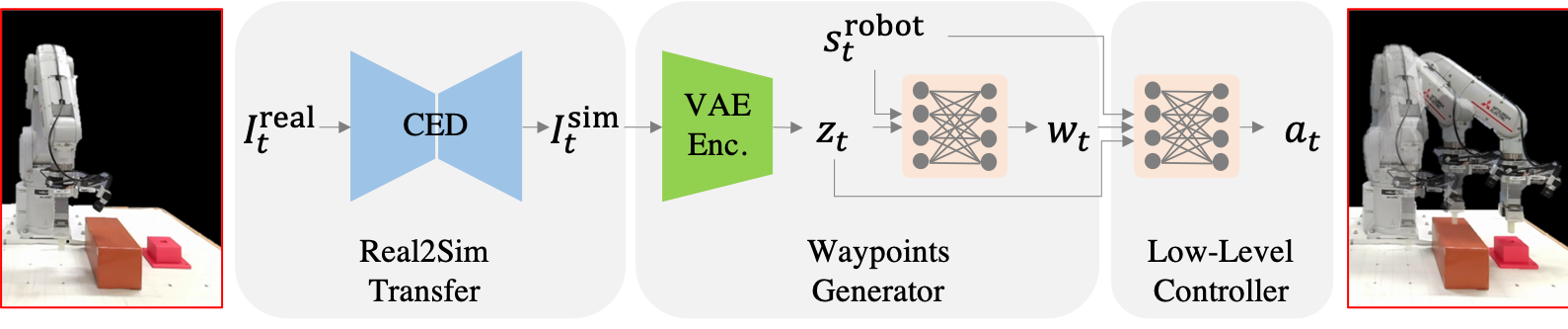}
        \caption{Architecture of the proposed method.
        The whole architecture consists of three modules: a Real2Sim transfer module converts the real depth image to the simulated one using a Convolutional Encoder-Decoder (CED). The simulated images are then compressed to latent variables by using a Variational Auto Encoder (VAE), and the waypoints generator takes the latent variables of the simulated image and robot's state, and generates the waypoints that roughly guide the agent to the desired goal. The input to the RL agent is a concatenation of the waypoints, the latent variables, and the internal state of robot, and outputs a low-level action to control the robot.}
        \label{fig:concept}
    \end{figure}

 The main idea of the proposed method is to encode a robot trajectory by a fixed number of waypoints in Cartesian or joint space, to be followed in a set order, and have a deep neural network produce the correct waypoints for a novel goal state and/or changed environment. To this end, the goal state and current environment are presented as inputs to a convolutional neural network (CNN), in the form of an image. Correspondingly, the coordinates (Cartesian or joint angles) of the  waypoints are presented as outputs of the CNN. A training set for the CNN is produced by running an offline planner such as RRT, for a large number of environments, respectively input images. The CNN is trained over this data set, and upon completion of training, is used to produce suitable waypoints for novel inputs that are observed in novel environments. The trajectory defined by the waypoints is then followed by a low-level action controller obtained by means of RL. The architecture of the end-to-end training method is shown in \fref{fig:concept}.    


\subsection{Waypoints Generator}\label{subsec:waypoints_generator}
    Learning an optimal policy from high-dimensional inputs, such as images, generally requires a huge number of interactions in the environment, and also bigger neural networks, and is thus not very suitable for using the policy on a real system.
    Although learning to produce an optimal action needs to solve an optimization problem, and thus requires significant computational time, generating an optimal path (i.e., the sequence of waypoints) is relatively easy using a prior path planning method, and using DNNs in a supervised learning mode is also easy to implement.
    Those waypoints can then be used to guide the agent to a goal position~
    \cite{ota2019trajectory,ota2020efficient}.
    We use this idea to first train a module for the agent that can predict an optimal path using information about an environment encoded via a depth image and the robot's state.

    We follow the same procedure as in~\cite{ota2020efficient}: we first generate a data set that consists of pairs of depth images in a simulator $I^\text{sim}$, and optimal paths $\bm{w}$, which are generated by using Bi-directional RRT$^\ast$~\cite{jordan2013optimal} with a random start, goal, and obstacle position. Then, the waypoints generator learns to output the waypoints by minimizing the following objective function:
    \begin{equation}
    	\mathcal{L}_\text{waypoints} =  \mathbb{E}_{I^\text{sim} \sim p_{\text {sim}}}\left[ \| G(I^\text{sim}, \bm{s}^\text{robot}) - \bm{w} \|^2_2 \right] \label{eq:waypoints}.
    \end{equation}

    The waypoints generator, $G$ in \eref{eq:waypoints}, consists of a CNN that takes a depth image $I^\text{sim}$ and a robot's state $\bm{s}^\text{robot}$, and generates a short horizon path $\bm{w}$ that consists of $5$ waypoints, each of which has $3$- or $6$-dimensional relative position or angle with respect to the current state of the agent. Thus, the SL module provides the agent with a rough, collision-free path that it can follow to reach the desired goal state.

\subsection{Path-Conditioned RL} \label{subsec:path_conditioned_rl}
    We utilize the short horizon optimal paths, which are generated by the waypoints generator, to improve the sample efficiency of training an RL agent and its generalization capability over novel environments.
    In order to do that, we exploit the waypoints in two ways: reward shaping and curriculum learning.

\paragraph{Reward Shaping}
    Since the depth image of the environment does not directly provide an agent with the information to accomplish a task, it generally needs a huge amount of interaction with environments to learn an optimal policy.
    We mitigate this problem by utilizing waypoints.
    Although the waypoints do not have dynamical information that can directly control a robot, it can roughly guide the agent to move to the goal state by using the waypoints for reward shaping.
    We follow a formulation similar to the one proposed in~\citep{ota2019trajectory,ota2020efficient}, which we call \textit{path-conditioned RL}.
    It defines a reward function by combining the original RL formulation and reference path (waypoints) $\bm{w}$ based function as
    \begin{equation}
    r(\bm{s}_t, \bm{a}_t, \bm{w}_t) = f(\bm{s}_t, \bm{a}_t) + h(\bm{s}_t, \bm{a}_t, \bm{w}_t), \label{eq:path_conditioned_rl}
    \end{equation}
    where $f(\bm{s}_t, \bm{a}_t)$ comes from the original RL formulation, and $h(\bm{s}_t, \bm{a}_t, \bm{w}_t)$ is waypoints based function. We define $f(\bm{s}_t, \bm{a}_t)$ as:
    \begin{equation}
        f(\bm{s}_t, \bm{a}_t) = \lambda_1 \mathbb{I}_{\rm collision} + \lambda_2 \mathbb{I}_{\rm goal} + \lambda_3 \mathbb{\| \ddot{\bm{\theta}} \|}, \label{eq:reward_orig}
    \end{equation}
    where $\lambda_1 \mathbb{I}_{\rm collision}$ penalizes collision with obstacles, and $\lambda_2 \mathbb{I}_{\rm goal}$ encourage the agent to achieve a reach task, and $\lambda_3 \mathbb{\| \ddot{\bm{\theta}} \|}$ penalizes angular accelerations to generate a smoother trajectory.
    We then define $h(\bm{s}_t, \bm{a}_t, \bm{w}_t)$ as
    \begin{equation}
        h(\bm{s}_t, \bm{a}_t,\bm{w}_t) = \lambda_4 d_{\rm path} + \lambda_5 n_{\rm progress},
    \label{eq:reward_way_points}
    \end{equation}
    where $d_{\rm path}$ is the distance to the reference path (waypoints) and $n_{\rm progress}$ is the progress along the path.
    The first term limits the exploration area by penalizing the distance to the waypoints, whereas the second term encourages the agent to move along the waypoints towards the goal state.
    More details can be found in~\cite{ota2020efficient} and the Appendix~\ref{subsec:app_rl}.

\paragraph{Curriculum Learning}

It is important to note that the waypoint generator is only based on kinematic planning for the robot in the given environment. Consequently, it is possible that the RL agent starts deviating from the sequence of waypoints as it tries to follow these waypoints using feasible control actions, and thus may start diverging if the episode length is long. This could degrade (or even destabilize) the whole training process as it can guide the waypoint generator to domains where it does not generalize well. In these situations, the RL agent cannot achieve the original task of reaching a goal state because the waypoints generator is not able to accurately guide the agent to the goal state. To make sure that the RL agent stays close to the sequence of waypoints generated by the waypoint generator, we initialize the robot state to waypoints which are closer to the goal state of the robot, instead of starting at the initial state of the robot. We iterate over a sequence of such random initialization of RL episodes which makes RL training easier. This allows the RL agent to collect informative samples from different parts of the trajectory and thus allows stable learning (as the RL agent does not diverge from the planned sequence of waypoints). 
    

\subsection{Real2Sim Transfer} \label{subsec:sim2real}
    In general, training a deep RL agent could be very sample inefficient, and we would like to minimize data collection on the real system when training the RL agent.
    In order to do that, we train a style transfer model that maps images taken in the real system to a simulator using a Convolutional Encoder-Decoder (CED).
    The CED learns a Real2Sim generative model $F:I^\text{real} \rightarrow I^\text{sim}$ by minimizing the following objective function:
    \begin{equation}
    	\mathcal{L}_{\text{real2sim}}(F)=\mathbb{E}_{I^\text{real} \sim p_{\text {real}}}\left[\|F(I^\text{real})-I^\text{sim}\|_{1}\right] . \label{eq:real2sim}
    \end{equation}
    Our aim is to utilize the Real2Sim generator $F$ to convert the real images $I^\text{real}$ to simulated ones $I^\text{sim}$, so that we can optimize the policy only in the simulator.
    To do so, we first generate pairs of simulated depth images and waypoints as described in  \sref{subsec:waypoints_generator}, and reproduce the same scenario in the real system: locate the obstacles and the goal to the same location, and control the robot to imitate the angles to make the pose look similar in simulation.
    Then, we train the CED by minimizing the objective function of \eqref{eq:real2sim}.





    \section{Experimental Settings}
This section briefly describes the experimental settings including the environments, hardware, and tasks used in experiments to validate the proposed method (more details in Appendix). 
\subsection{MDP} \label{subsec:mdp}
\paragraph{States}
	The states of the system consist of current joint angles $\bm{\theta}_t \in \mathbb{R}^{6}$, angular velocities $\dot{\bm{\theta}}_t \in \mathbb{R}^{6}$ in configuration space, and waypoints $\bm{w}_t \in \mathbb{R}^{5\times 6}$ that lead the agent to the goal state.
	Since the robot state and waypoints do not contain spatial information about an environment, we also add latent variables $\bm{z}_t \in \mathbb{R}^{64}$ of the environment to the state, which is produced by using Variational Auto Encoder (VAE)~\cite{kingma2013autoencoding}, instead of using raw pixels to make the problem easier to solve~\cite{yarats2019improving,kendall2019learning}.
	If the agent moves in Cartesian space, then we use the position and velocities of the end tool of the robot, instead of using joint angles. See \fref{fig:concept} for more details of the network used for training.

\paragraph{Actions}
	The action of the agent $\bm{a}_t$ is the vector of angular velocities $\dot{\bm{\theta}}$ for the next step.
	We define a time step described as $\Delta t$.
	Therefore, the angles of the next step $\bm{\theta}_{t+1}$ can be calculated as 
	\begin{equation}
		\bm{\theta}_{t+1}=\bm{\theta}_t + \bm{a}_t \Delta t \label{eq:dynamics}.
	\end{equation}



\paragraph{Termination Condition}
	An episode terminates on the following three conditions: the joint angles of the agent $\bm{\theta}_t$ are sufficiently close to the goal state, or the number of steps of an episode is over a specified threshold, or the robot violates the known maximum joint angles.

\subsection{Hardware} \label{subsec:hardware}
	We use a MELFA RV-4FR robot, which is an industrial robot that has 6 degrees of freedom \cite{melfa}. We operate the robot in a position control mode where position commands are sent to the robot every $\Delta t=0.0035$ seconds, which is determined by the minimum operational time of the industrial robot we used in a real setting.
	The generated actions must ensure that joint angular accelerations are within a known specified range. Since lower acceleration is a desirable feature for a lot of industrial manipulators where direct torque control is not accessible, we penalize the angular acceleration when training our agent (see~\sref{subsec:path_conditioned_rl}).
    For the robot to perceive the environment, we use Azure Kinect, which is an inexpensive, commercial RGBD camera. Since the maximum frame rate of the camera is 30 FPS, which is much slower than the control step of the robot's controller, we implement a distributed system using ROS, so that the RL component does not need to wait for images.

\subsection{Tasks} \label{subsec:tasks}
    In order to evaluate the proposed method, we prepare two different tasks: \PegInsertion and \PickPlace.
    We use the MuJoCo~\cite{todorov2012mujoco} physics engine to simulate both of these tasks.

\begin{figure*}[t]
    	\begin{minipage}[]{0.66\linewidth}
    		\centering
    		\includegraphics[height=30truemm]{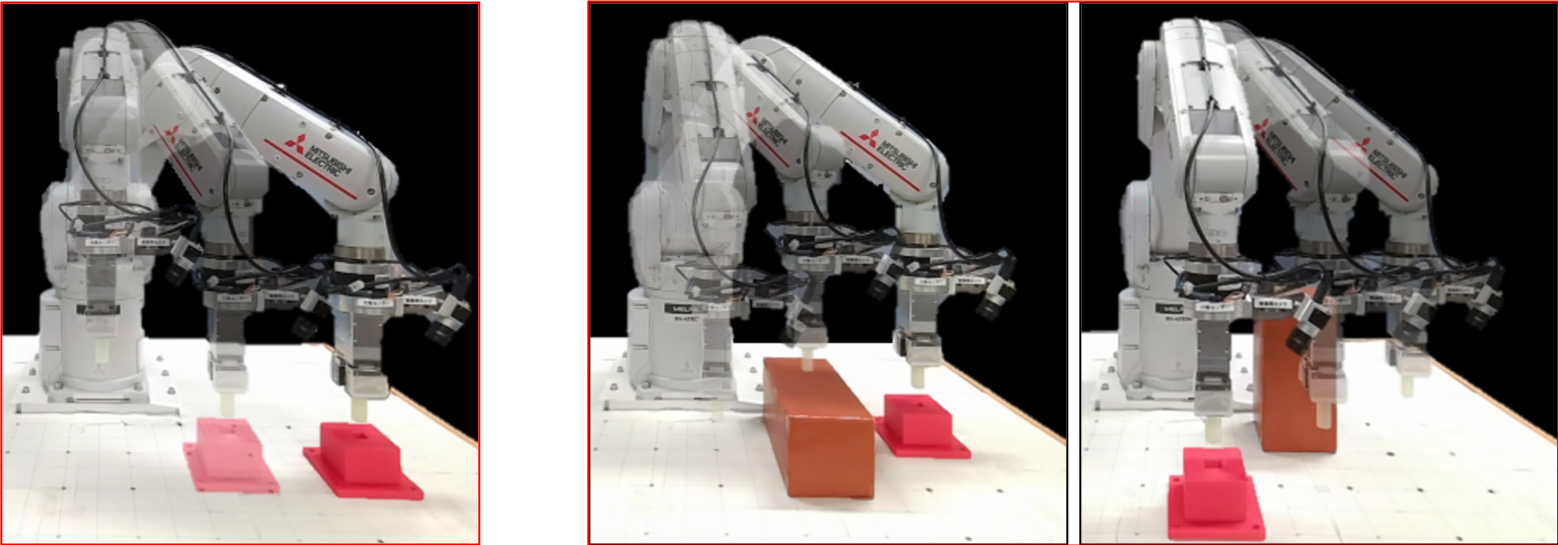}
    		\subcaption{\PegInsertion environment.}\label{fig:peg_insertion_env}
    	\end{minipage}
    	\begin{minipage}[]{0.33\linewidth}
    		\centering
    		\includegraphics[height=30truemm]{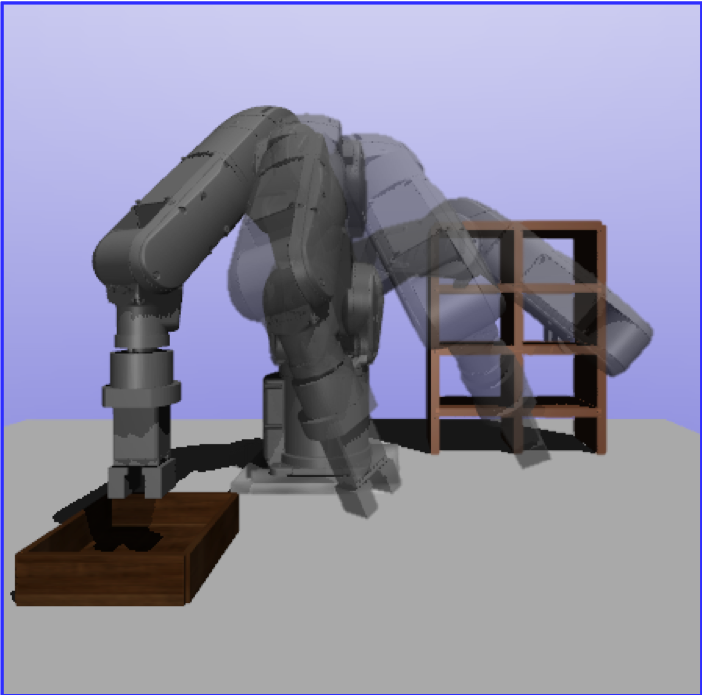}
    		\subcaption{\PickPlace environment.}\label{fig:pickplace_env}
    	\end{minipage}
    	\caption{The trajectories obtained by the proposed method for the two different settings.
    	    }
    	\label{fig:env_setting}
    \end{figure*}

    The \PegInsertion task is for simulating peg insertion, navigating a robotic hand which holds a white peg, and inserting it into a red hole located on top of the table, while avoiding collision with a number of $N^\text{obs}$ brown obstacles as illustrated in \fref{fig:peg_insertion_env}.
    Since the grasping and insertion is not the focus of this work, an episode starts from a random initial position with the robot holding the peg, with the goal of reaching a random hole position, without collision with randomly located obstacles.
    A video of the implementation of the algorithm on the real system is provided in the following link: \url{https://youtu.be/hE-Ew59GRPQ}.
    The generated trajectories must ensure that joint angles and angular velocities that describe the trajectories are within a known specified range. We, however, fix the orientation of the robot hand, since we assume the hole is located on top of the horizontal table. As a result, the control input is the velocity in Cartesian space.

    The \PickPlace task simulates a pick-and-place task, where the robot picks an object located in a two-row, three-column bookcase, and puts it into a box located on top of a table as depicted in \fref{fig:pickplace_env}. Each of the cubes in the bookshelf is $150$ mm deep, $150$ mm high and $200$ mm wide.
    The manipulator starts from an initial pose denoted by $\bm{\theta}_\text{start}$, and has to reach the box.
    The start position is sampled randomly from the center of each cube of the bookshelf, in order to imitate a grasping motion needed to grab an item.

    For both tasks, we determine the success of an episode based on the following conditions: 1) the path does not collide with obstacles, and 2) the robot arm reaches the goal position closely enough, specifically we define the threshold to be $d_\text{goal} = 50$ [mm], and 3) the episode length does not exceed a pre-defined length of $T=300$ steps.

    \section{Experimental Results}
In this section, we try to quantify generalization and performance of the proposed technique. 
Specifically, we try to answer the following questions:
\begin{itemize}
	\item What is the performance and generalization achieved by the waypoint generator?
	\item Can the end-to-end reactive motion planner generalize to novel environments?  
	\item Can the path-conditioned RL improve the sample efficiency and quality of the generated trajectories?
\end{itemize}

\subsection{Performance of the Waypoints Generator on the Real System} \label{subsec:waypoints_gen}
    To evaluate the quality of the waypoints generation module, we use two different metrics -- 1) success rate : success is declared in \sref{subsec:tasks} and 2) waypoint error : the average error in reproducing the waypoints generated by the Bi-directional RRT$^\ast$.


    As described in \sref{subsec:sim2real}, our aim is to train our policy in simulation, and we achieve this by learning a function that maps real images to simulated ones using CED to avoid exploration on the real system.
    Thus, we measure the success rate on different numbers of training data collected in the real system and compare the results of supervised learning of only real data (Real-Only) and our approach of learning the waypoints generator only in simulation (CED + Sim).
    We also tested other approaches, such as supervised learning of only simulated data, CycleGAN~\cite{zhu2017unpaired}, unsupervised domain adaptation~\cite{ganin2014unsupervised}, fine-tuning the simulation-based model with real data~\cite{richter2016playing}, and mixing the simulated and real images. We found that mixing the simulation and real data works best among these approaches. Therefore, we also compare it, denoting it as "Real + Sim" in the experiment. We generate $50k$ images in simulation, as it is relatively inexpensive to generate images in simulation. More details about the evaluation experiments are available in the Appendix~\ref{subsec:appendix_peg_insertion_exp}.


    \Tref{tab:results_way_points_gen} shows the comparison between different approaches using the two different metrics. We can clearly see that the combination of CED and waypoints generator trained in simulation successfully generalizes to a new environment, even if the training data decreases to 10\%. Thus, we can train policies in simulation with very little data required from the real system.

	\begin{table}[t]
		\caption{Success rate and average waypoints errors of the \PegInsertion task in the real system. The bold numbers show the highest success rate or minimum waypoints error. The percentages refer to percent of training data, not success rate or waypoints error.}
		\label{tab:results_way_points_gen}
		\begin{center}
			\begin{tabular}{lcccccccc} \toprule
			     & & \multicolumn{3}{c}{Success rate} & & \multicolumn{3}{c}{Waypoints error [mm]} \\
			     \cmidrule{3-5} \cmidrule{7-9}
			    & & 10\% & 50\% & 100\% & & 10\% & 50\% & 100\% \\ \midrule
			    \multirow{3}{*}{$N^\text{obs}=0$}
			    & Real-Only & 0.08 & 0.06 & 0.10 && 14.4 & 16.7 & 13.1 \\
			    & Real + Sim & $\bm{1.00}$ & $\bm{1.00}$ & $\bm{1.00}$ && $\bm{3.2}$ & $\bm{2.8}$ & $\bm{2.5}$ \\
			    & CED + Sim (Ours) & 0.80 & $\bm{1.00}$ & $\bm{1.00}$ && 5.7 & 3.3 & 3.4 \\ \midrule
			    \multirow{3}{*}{$N^\text{obs}=1$}
			    & Real-Only & 0.00 & 0.00 & 0.00 && 12.2 & 12.8 & 13.8 \\
			    & Real + Sim & 0.42 & 0.90 & 0.71 && 13.8 & $\bm{3.6}$ & 5.1\\
			    & CED + Sim (Ours) & $\bm{0.80}$ & $\bm{0.98}$ & $\bm{0.94}$ && $\bm{9.3}$ & 7.9 & $\bm{2.8}$ \\ 
			    \bottomrule
			\end{tabular}
		\end{center}
	\end{table}

\subsection{Generalization to Novel Environments}\label{subsec:results_generalization}
    Next, we evaluated the generalization capability of our method with respect to novel environments with our end-to-end policy, i.e., including a low-level action generation module using RL.
    As for the test environment, we prepared two different settings and tested over 50 trials on each setting in the real system.
    First, we tested generalization on a moving goal on the $N^\text{obs}=0$ environment, which changes its goal position during execution of the robot, before the robot has reached the previous position.
    Second, we changed the positions of the obstacles and the goal during execution by stopping the robot in order to ensure the safety of the operator.
    The definition of success is whether the agent reaches within $d_{\text{goal}}$ of the goal position.
    We compared our method of CED + Sim against the Real-Only model, both of which use 100\% of data as in the previous experiment.

    \Tref{table:goal_param_success_rate} shows the result of the experiments.
    It suggests our method can reactively track the moving goal positions, as depicted in the \fref{fig:peg_insertion_env}, and also reactively change the path with more complex settings of changing both obstacle and goal positions and/or orientation during an episode.
    Videos that show this behavior on the real system are provided in the \url{https://youtu.be/hE-Ew59GRPQ}.
	\begin{table*}[t]
		\caption{Performance on generalization task in the real system. The results are averaged over $50$ trials.}
		\label{table:goal_param_success_rate}
		\begin{center}
			\begin{tabular}{ccccc} \toprule
			    & \multicolumn{2}{c}{Success rate} & \\
			      & Random obstacles & Moving goal \\ \midrule
			     Real-Only & 0.00 & 0.00 \\
			     CED + Sim (Ours) & $\bm{0.90}$ & $\bm{1.00}$  \\
			     \bottomrule
			\end{tabular}
		\end{center}
	\end{table*}

\subsection{Improving Sample Efficiency and Final Performance using Path-Conditioned RL} \label{subsec:rl_performance}
    As described in \sref{subsec:path_conditioned_rl}, we believe that path-conditioned RL lets the agent explore more efficiently, and thus leads to better sample efficiency. Moreover, we think the RL-based low-level controller can generate optimal trajectories in terms of minimizing angular accelerations and time to reach a goal state.
    To verify these, we conducted the following experiments on the \PickPlace environment in simulation.

\paragraph{Sample Efficiency}
    We tested whether the proposed method can improve the sample efficiency compared with a baseline RL agent, which does not contain waypoints in the state.
    Therefore, the state of the baseline agent will be the concatenation of the latent variables $\bm{z}_t$ obtained by using VAE, and robot's state $\bm{s}_t^\text{robot}$, and trained only from the reward function that removes waypoints-related terms, i.e., removed $h(\bm{s}_t, \bm{a}_t, \bm{w}_t)$ from Eq.~\eqref{eq:path_conditioned_rl}.
    Both agents were trained with the Soft Actor Critic (SAC) algorithm~\cite{haarnoja2018soft} with the same hyperparameters as in the paper.
    For a fair comparison, we evaluated both agents with the same reward function as the baseline agent. 

\begin{figure*}[t]
    	\begin{minipage}[]{0.33\linewidth}
    		\centering
    		\includegraphics[width=\columnwidth]{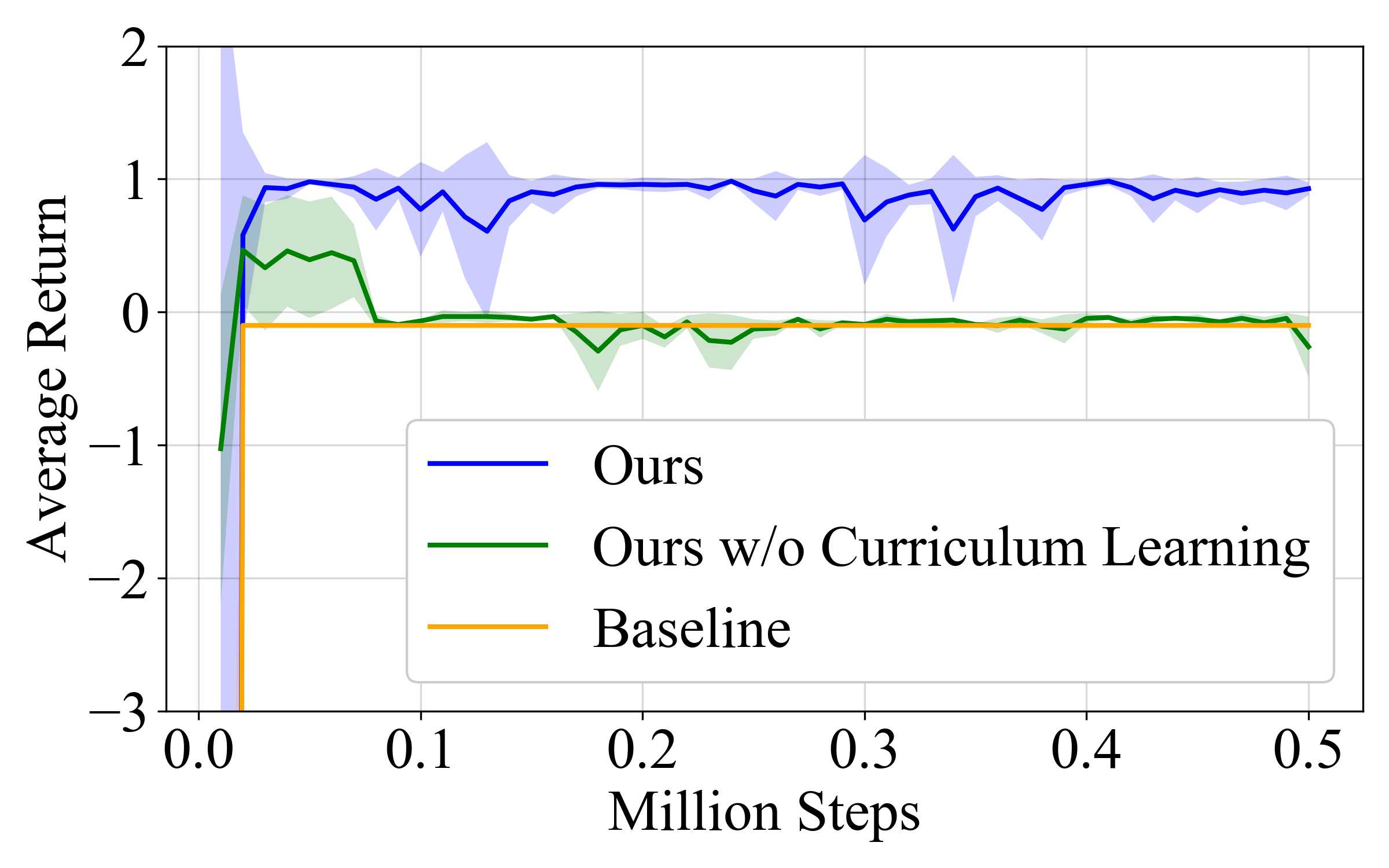}
    		\subcaption{Episodic return}\label{fig:results_return}
    	\end{minipage}
    	\begin{minipage}[]{0.33\linewidth}
    		\centering
    		\includegraphics[width=\columnwidth]{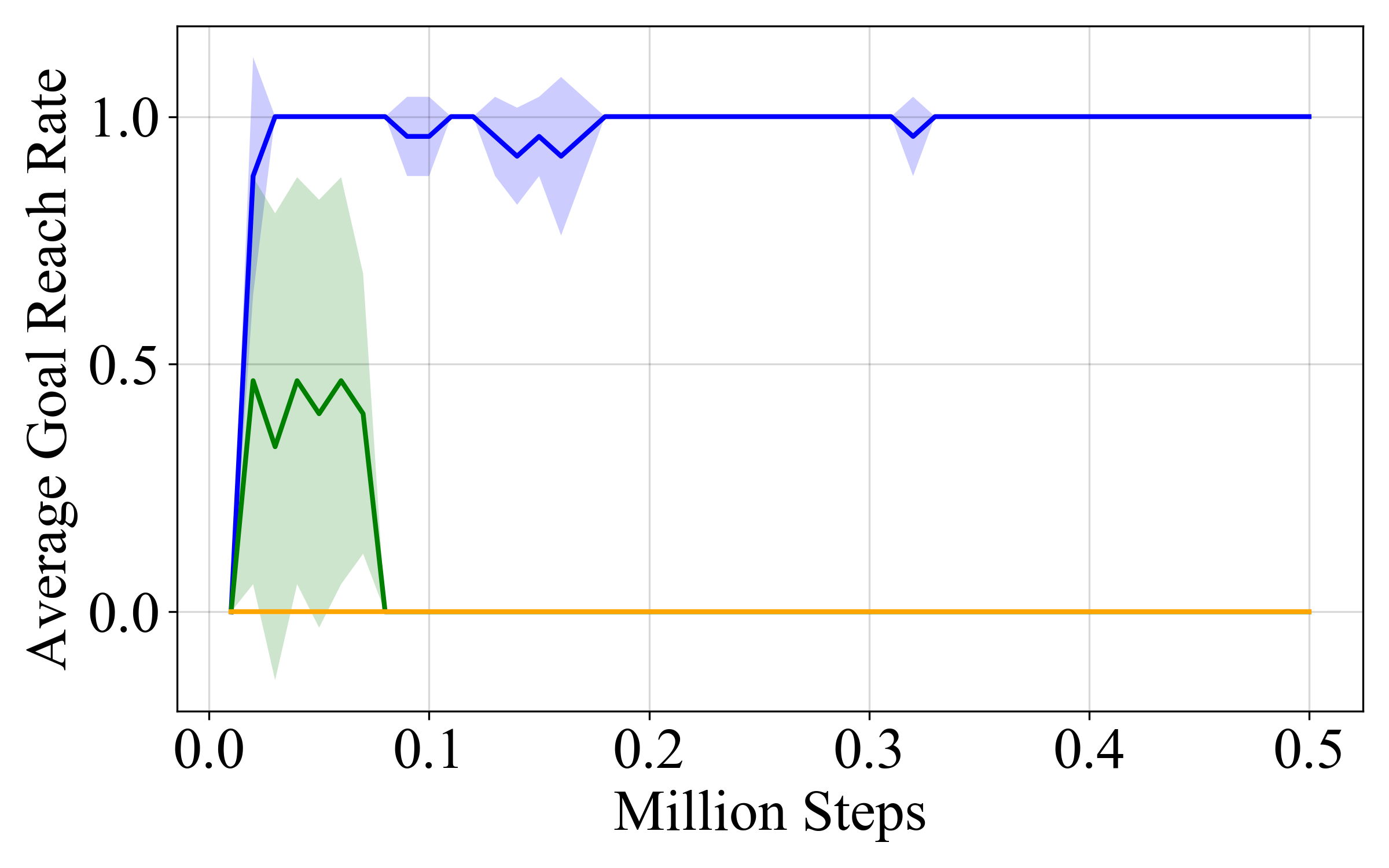}
    		\subcaption{Goal reach rate}\label{fig:results_goal_rate}
    	\end{minipage}
    	\begin{minipage}[]{0.33\linewidth}
    		\centering
    		\includegraphics[width=\columnwidth]{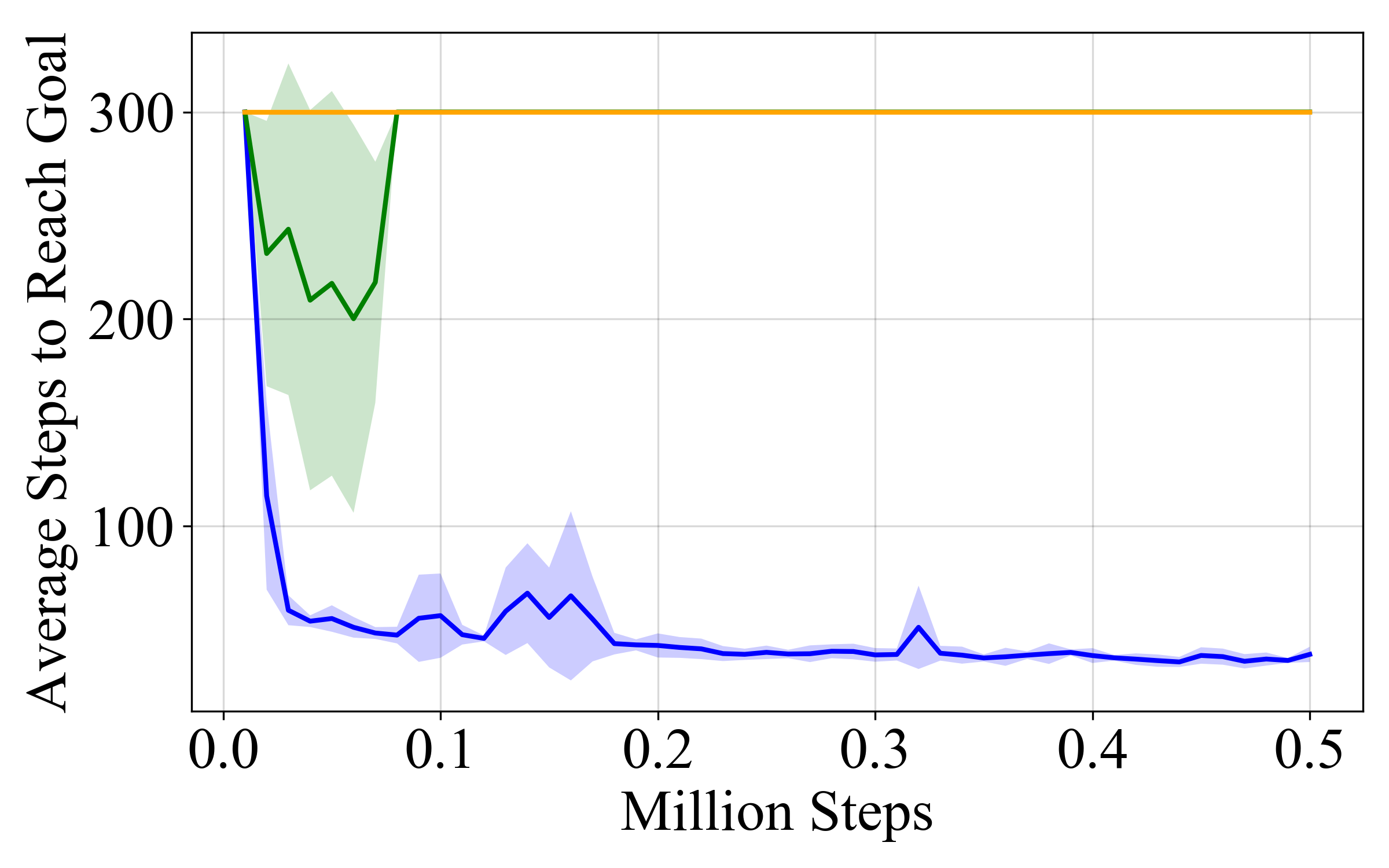}
    		\subcaption{Steps to Reach Goal}\label{fig:results_steps}
    	\end{minipage}
    	\caption{Training curves on each training type. The solid lines represent average returns over five instances with different random seeds. The shaded region represents the standard deviation of the five instances. Our approach outperforms baseline on both sample efficiency and final performance.}
    	\label{fig:rl_results}
    \end{figure*}

    \Fref{fig:rl_results} demonstrates the training curves of resulting episodic returns, goal reach rate, and number of steps needed to reach a goal state.
    Comparing our method with the baseline, we see that solving the reaching task only from robot's state and latent variables of the depth image is difficult, and the use of the waypoints can make the original task substantially easier. It also improves the sample efficiency and achieves better final performance.
    Next, we compared the result of our path-conditioned RL with and without curriculum learning. The result suggests that just using waypoints does not successfully converge to achieve the desired task, but curriculum learning plays an important role to achieve the task. We believe that using curriculum learning makes training more stable, thus leading to faster convergence (see \sref{subsec:path_conditioned_rl}).


\paragraph{Quality of Generated Trajectories}
    Finally, we evaluated the quality of the generated trajectories in terms of average angular accelerations during an episode, and time to reach a goal state.
    We compared our method against a combination of Bi-directional RRT$^\ast$ and a PID controller to follow the generated path as a baseline.
    We use the PID controller in the officially provided simulator of the robot used in the real experiment~\cite{rttoolbox}.
    We evaluated 6 different scenarios, each of which starts from imitating to pick an object in a cube (1 to 6 corresponds to from top left to bottom right in \fref{fig:pickplace_env}), and reach a box located in the same position for fair comparison.

    \Tref{tab:results_quality} shows the results.
    It shows that the use of low-level RL-based controller does improve the quality of the generated path in terms of both time to reach the goal state and in minimizing angular accelerations during an episode.

	\begin{table}[t]
		\caption{Time [sec] to reach goal, and average angular acceleration $\text{[rad/sec}^2\text{]}$ during execution.}
		\label{tab:results_quality}
		\begin{center}
			\begin{tabular}{ccccccccc} \toprule
				& & \multicolumn{6}{c}{\PickPlace} \\
				& & 1 & 2 & 3 & 4 & 5 & 6 \\ \midrule
				\multirow{2}{*}{Time [sec]} &
				Baseline & 0.89 & 0.92 & 0.80 & 0.96 & 1.25 & 1.26 \\
				& Ours & \bf{0.48} & \bf{0.51} & \bf{0.59} & \bf{0.52} & \bf{0.57} & \bf{0.54} \\ \midrule
				Angular Accelerations &
				Baseline & 696 & 672 & 687 & 702 & 4365 & 4539 \\
			    $\text{[rad/sec}^2\text{]}$ & 
			    Ours & \bf{124} & \bf{125} & \bf{163} & \bf{135} & \bf{153} & \bf{136} \\ \bottomrule
			\end{tabular}
		\end{center}
	\end{table}

    \section{Conclusion}
In this paper, we presented an end-to-end policy that can perform reactive planning for robotic systems in dynamic environments. The proposed method uses kinematic planning, supervised learning, and RL to train an agent that can do reactive planning using input from a depth camera. To minimize training with the real system, we use a real2sim transfer function, and train the agent entirely in simulation with very few real data samples. We show that the proposed algorithm outperforms several baseline methods, and allows a $6$-DoF manipulator arm to perform reactive planning in complex dynamic environments in an end-to-end fashion.
    \bibliography{references}

    \appendix
    \section{Experimental Details}

\subsection{Real System}
In this section, we provide more details of the real system which we use for experiments in \sref{subsec:waypoints_gen} and \sref{subsec:results_generalization}.


\paragraph{Depth Sensor}
As described in \sref{subsec:hardware}, we use Azure Kinect for the real experiments, and reproduce the same setup in the simulator using MuJoCo.
For the MuJoCo simulator, the raw pixel values $I^\text{sim-raw}$ are first converted to real distance values. Referring to~\cite{dmcontrol}, the conversion can be computed as:
\begin{equation}
    I^\text{sim-dist}_{i,j} = p^\text{near} / (1 - I^\text{sim-raw}_{i,j} \times (1 - p^\text{near} / p^\text{far})),
\end{equation}
where $I^\text{sim-raw} \in [0, 1]^{W \times H}$ is the depth image whose width and height is $W$ and $H$, and the $p^\text{near}$ and $p^\text{far}$ are camera parameters.
Then, we clip only regions of interest distance $d_\text{min}, d_\text{max}$ as:
\begin{equation}
    I^\text{sim}_{i,j} = \max(\min(I^\text{sim-dist}_{i,k}, d_\text{max}), d_\text{min}),
\end{equation}
where we specifically use $(d_\text{min}, d_\text{max}) = (0.5, 1.5)$ [m] for our setting.
Finally, the pixel values are normalized so that the minimum value becomes $0$ and the maximum value becomes $1$.

\paragraph{Obstacles}
We use the same shape obstacle over all experiments whose size is $(W \times H \times D) = (122, 248, 114)$ [mm].

\paragraph{Peg and Hole}
The peg and hole that we used for the real experiments are printed using a 3D-printer with the diameter of $\phi^\text{peg} = 20$ [mm], and $\phi^\text{hole} = 21$ [mm].

\subsection{Peg Insertion Experiments} \label{subsec:appendix_peg_insertion_exp}

\paragraph{Robot's State Space}
The start, goal, obstacle's positions are randomly sampled from a position defined in \Tref{tab:randomization_area_peg_insertion} when an episode starts.
The randomization area for generating an optimal path using Bi-directional RRT$^\ast$ is also defined in \Tref{tab:randomization_area_peg_insertion}.

\begin{table}[htbp]
    \centering
    \caption{The randomization area from which start, goal, obstacle positions, and robot state are sampled in the \PegInsertion task. Note that the orientation of the obstacles have three choices (rotate $\pi/2$ along $Y$ or $Z$ axes), and the height ($Z$-axis value) of the obstacles and the goal does not change due to a geometrical constraint.}
        \begin{tabular}{cc|rr} \toprule
        Randomization Target & Axis & Min. [m] & Max. [m] \\ \midrule 
        \multirow{3}{*}{Start, Goal, Obstacle, Robot's state}
        & $X$ & 0.20 & 0.40 \\
        & $Y$ & -0.40 & 0.40 \\
        & $Z$ & 0.05 & 0.30 \\
         \bottomrule
    \end{tabular}
    \label{tab:randomization_area_peg_insertion}
\end{table}

\paragraph{Settings for the Experiment in \sref{subsec:waypoints_gen}}
For the experiment in \sref{subsec:waypoints_gen}, where we evaluate the performance of the waypoints generator with respect to the number of data collected in the real system, we carefully divide the dataset into training and evaluation while not allowing the each data to overlap each other.
In order to do that, we sample goal positions for each dataset from discrete positions of a grid of $9 \times 3=27$ with each grid size is $100 \times 100$ [mm], and results in the total grid size is $200 \times 800$ [mm].
From the grid, we assigned $22/27$ of goal positions for training, and the others for evaluation so that the distribution of the goal positions in the two different dataset do not overlap. Furthermore, we divided the training dataset of $22/27$ goal positions into three subsets, which roughly contains 10\%, 50\%, 100\% of data sampled from different goal positions again, whose data size is roughly 10K data for 100\% dataset. It is noted that such a discrete grid is only created for evaluation purposes.

Since the experiment in \sref{subsec:waypoints_gen} does not include subsequent low-level action module to evaluate the waypoints generator, we move the robot to the closest waypoint by sending a position command to the real system (without using the RL trained policy).

\begin{figure}[t]
    \centering
    \includegraphics[width=\textwidth]{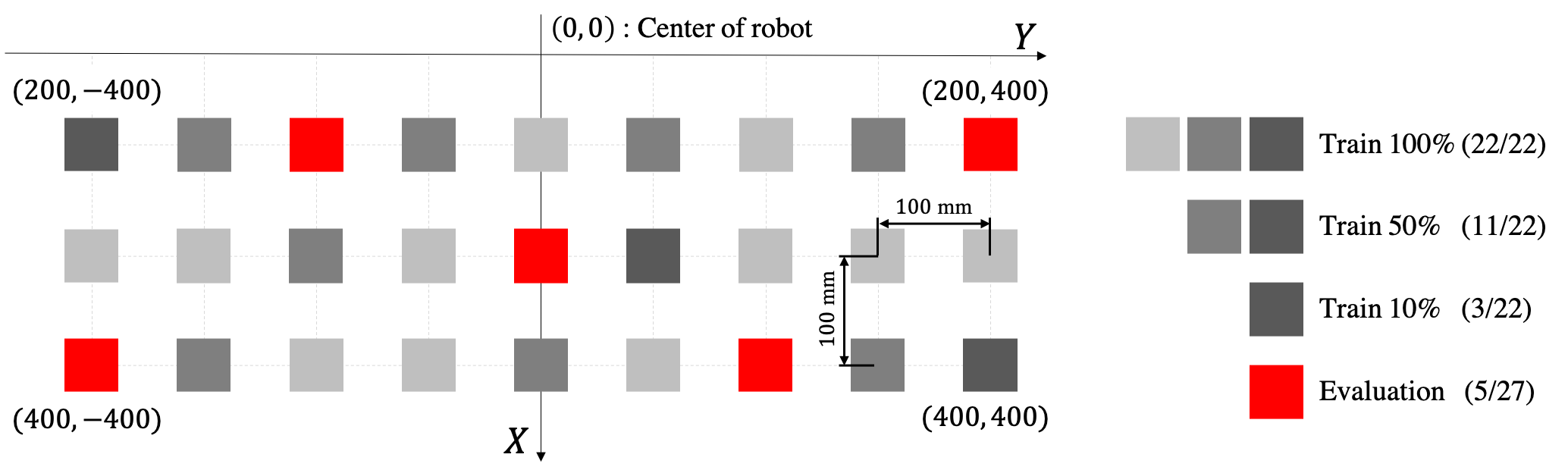}
    \caption{Goal positions [mm] where each training and evaluation data are sampled. The position is sampled from a grid of $9 \times 3=27$ with each grid size is $100 \times 100$ [mm].}
    \label{fig:randomization_area_peg_insertion}
\end{figure}

\paragraph{Settings for the Experiment in \sref{subsec:results_generalization}}
As for the CDE model used in the generalization experiments in \sref{subsec:results_generalization}, we use the same model with the one trained using 100\% as defined in the previous section.

The distribution of the goal position, however, is different from the previous experiment: the goal is sampled continuously from the randomization area defined in \Tref{tab:randomization_area_peg_insertion}, i.e., not categorically from the grid in \fref{fig:randomization_area_peg_insertion}.

\subsection{Pick-and-Place Experiments}
\paragraph{Robot's State Space}
The goal position is randomly sampled from a range defined in \Tref{tab:randomization_area_pick_place} when an episode starts.
The start position is randomly sampled from categories of 6 different cube location, where 1 to 6 corresponds to from top left to bottom right in \fref{fig:pickplace_env} as described in \sref{subsec:rl_performance}.
The randomization area for generating an optimal path using Bi-directional RRT$^\ast$ is also defined in \Tref{tab:randomization_area_peg_insertion}.

\paragraph{Settings for the Experiment in \sref{subsec:rl_performance}}
In order to compare the generated trajectories of ours against the PID controller in the officially provided simulator, we fix a goal position to be $\bm{x}^\text{goal}=(0.4, -0.2)$.

\begin{table}[t]
    \centering
    \caption{The randomization area from which start, goal, and robot's state are sampled. Note that the height ($Z$-axis value) of the goal does not change due to a geometrical constraint.}
        \begin{tabular}{cc|rr} \toprule
        Randomization Target & Axis & Min. [deg] & Max. [deg] \\ \midrule 
        \multirow{6}{*}{Robot's state}
        & $\theta_1$ & -240 & 240 \\
        & $\theta_2$ & -120 & 120 \\
        & $\theta_3$ & 0 & 164 \\
        & $\theta_4$ & -200 & 200 \\
        & $\theta_5$ & -120 & 120 \\
        & $\theta_6$ & -360 & 360 \\ \midrule
        \multirow{2}{*}{Goal}
        & $X$ & 0.40 & 0.45 \\
        & $Y$ & -0.20 & 0.30 \\
        \bottomrule
    \end{tabular}
    \label{tab:randomization_area_pick_place}
\end{table}
\section{Training Details}
As depicted in \fref{fig:concept}, we train four neural networks models to accomplish the task of reactive planning in the real system: 1) the Convolutional Encoder-Decoder (CDE) for Real2Sim transfer, 2) the VAE model that extracts latent variables of an environment, 3) the waypoints generator which generates waypoints $\bm{w}$ that roughly guides the agent to a goal state, and 4) the RL-based low-level controller that produces an optimal action in terms of time to reach goal and generating a smoother trajectory.
The four modules are trained separately, and we train each module while freezing the parameters of the other modules.
In this section, we provide details about these models.

\subsection{Real2Sim Transfer}
We adopt our architecture of the Real2Sim transfer CDE model $F:I^\text{real} \rightarrow I^\text{sim}$ from \cite{Johnson2016Perceptual} referring to \cite{zhu2017unpaired} with the 2 residual blocks instead of 9 for speeding up computation.
We train the CDE model by minimizing Eq.~\eqref{eq:real2sim} with Adam optimizer for 1000 epochs with a learning rate of $0.0002$ as in \cite{zhu2017unpaired}.

\subsection{VAE}
The VAE model we train for extracting latent variables $\bm{z}_t$ of an environment consists of a convolutional encoder-decoder network, i.e., a convolutional encoder $g_\phi$ maps a depth image of the simulator $I_t^\text{sim}$ to a low-dimensional latent variables $\bm{z}_t$, and a deconvolutional decoder $f_\theta$ reconstructs $\bm{z}_t$ back to the original image $I_t^\text{sim}$.
\Tref{tab:vae_architecture} show the architecture of the VAE.

The VAE model is trained by maximizing the following objective:
\begin{equation}
    J_\mathrm{VAE} =\mathbb{E}_{I_t^\text{sim} \sim p^\text{sim}} \left[\mathbb{E}_{\mathbf{z}_{t} \sim q_{\phi}\left(\mathbf{z}_{t} \mid I_t^\text{sim}\right)}\left[\log p_{\theta}\left(I_t^\text{sim} \mid \mathbf{z}_{t}\right)\right]\right.
\left.-\beta D_{\mathrm{KL}}\left(q_{\phi}\left(\mathbf{z}_{t} \mid I_t^\text{sim}\right) \| p\left(\mathbf{z}_{t}\right)\right)\right]
\end{equation}
with the $50K$ size of dataset collected in the simulator as described in \sref{subsec:waypoints_gen}, and Adam optimizer for 1000 epochs with a learning rate of $0.0001$.
For more details of VAE, interested readers are referred to~\cite{kingma2013autoencoding}.

\begin{table}[t]
    \centering
    \caption{Architecture of VAE.}
    \begin{tabular}{ccccc}
        \toprule
        & Shape & Layer size & Stride & Type \\ \midrule
        \multirow{6}{*}{Encoder}
        & $64 \times 64 \times 1$ & - & - & Input: depth image $I^\text{sim}$ \\
        & $32 \times 32 \times 16$ & $3 \times 3 \times 16$ & 2 & Convolution + ReLU \\
        & $16 \times 16 \times 32$ & $3 \times 3 \times 32$ & 2 & Convolution + ReLU \\
        & $8 \times 8 \times 64$ & $3 \times 3 \times 64$ & 2 & Convolution + ReLU \\
        & $4096$ & - & - & Flatten \\
        & $64 \times 2$ & - & - & Fully Connected: output means and variances \\
        \midrule
        \multirow{7}{*}{Decoder}
        & $64$ & - & - & Latent variables $\bm{z}$ \\
        & $4096$ & - & - & Fully Connected + ReLU \\
        & $8 \times 8 \times 64$ & - & - & Reshape \\
        & $16 \times 16 \times 64$ & $3 \times 3 \times 64$ & 2 & Deconvolution + ReLU \\
        & $32 \times 32 \times 32$ & $3 \times 3 \times 32$ & 2 & Deconvolution + ReLU \\
        & $64 \times 64 \times 16$ & $3 \times 3 \times 16$ & 2 & Deconvolution + ReLU \\
        & $64 \times 64 \times 1$ & $3 \times 3 \times 1$ & 1 & Convolution: output reconstructed image $\hat{I}^\text{sim}$ \\ \bottomrule
    \end{tabular}
    \label{tab:vae_architecture}
\end{table}

\subsection{Waypoints Generator}
As described in \sref{subsec:waypoints_generator}, the waypoints generator consists of a CNN that takes a depth image $I^\text{sim}$ and a robot's state $\bm{s}^\text{robot}$, and generates a short horizon path $\bm{w}$ that consists of $N^\text{waypoints} = 5$ waypoints, each of which has $3$- or $6$-dimensional relative position or angle with respect to the current state of the agent.
\Tref{tab:waypoints_cnn_architecture} shows the architecture of the waypoints generator. 
The depth image is converted into latent variables using the VAE model described in the previous section, and the concatenation of the latent variables $\bm{z}$ and robot's state $\bm{s}^\text{robot}$ is inputted into the waypoints generator model to produce the waypoints.

In order to train the model, we first collect $50K$ pairs of depth images, robot's states, and waypoints generated by using Bi-directional RRT$^\ast$ with random start, goal, and obstacles position.
We then train the model with Adam optimizer for 500 epochs with a learning rate of $0.001$.

\begin{table}[t]
    \centering
    \caption{Architecture of the waypoints generator. The $N^\text{state}=3$ if the robot moves in Cartesian space, or $N^\text{state}=6$ if in configuration space.}
    \begin{tabular}{cccc}
        \toprule
        Shape & Layer size & Stride & Type \\ \midrule
        $64 + N^\text{state}$ & - & - & Concatenate latent variables $\bm{z}$ and robot's state $\bm{s}^\text{robot}$\\
        $256$ & - & - & Fully Connected + ReLU \\
        $N^\text{waypoints} \times N^\text{state}$ & - & - & Fully Connected: output waypoints $\bm{w}$ \\
        \bottomrule
    \end{tabular}
    \label{tab:waypoints_cnn_architecture}
\end{table}


\subsection{Reinforcement Learning} \label{subsec:app_rl}
This section summarizes extra details of the reward function and the curriculum learning setting we used in \sref{subsec:rl_performance}.

\paragraph{Reward}
\Tref{tab:rewards} summarizes the coefficients of the reward terms defined in \eqref{eq:reward_orig}, and \eqref{eq:reward_way_points}.

\begin{table}[t]
    \centering
    \caption{Coefficients of each reward term $\lambda_i$ used in our experiments.}
    \begin{tabular}{lrl} \toprule
        Term & Value & Description \\ \midrule
         $\lambda_1 \mathbb{I}_{\rm collision}$ & $-0.1$ & Obstacle collision penalty \\
         $\lambda_2 \mathbb{I}_{\rm goal}$ & $1.0$ & Goal reach reward \\
         $\lambda_3 \mathbb{\| \ddot{\bm{\theta}} \|}$ & $-0.0001$ & Angular acceleration penalty \\ 
         $\lambda_4 d_{\rm path}$ & $0.1$ & Distance penalty to the closest waypoints \\
         $\lambda_5 n_{\rm progress}$ & $-1.0$ & Distance reward going toward the goal state \\
         \bottomrule
    \end{tabular}
    \label{tab:rewards}
\end{table}

\paragraph{Curriculum Learning}
Here we explain the curriculum learning setting we used in \sref{subsec:rl_performance}.
First, we randomly sample the number of steps $N^\text{waypoints-step}$, which specifies how many steps we rollout using the waypoints generators from an initial state.
Then, we iteratively produce the waypoints for $N^\text{waypoints-step}$-steps by making use of the waypoints $\bm{w}_t$ to evolves the environment at each time steps $t$.
More specifically, we compute the next joint angles of the robot as:
\begin{equation}
    \bm{\theta}_{t+1} = \bm{\theta}_t + \bm{w}^\text{closest}_t,
\end{equation}
where the $\bm{w}^\text{closest}\in\mathbb{R}^6$ is the closest waypoints, and then we set the robot's state to the internal property of the simulator.
Algorithms~\ref{alg:curriculum_learning_procedure} shows the procedure of the curriculum learning setting we used in .
	\begin{algorithm}
		\caption{Curriculum Learning }\label{alg:curriculum_learning_procedure}
		\begin{algorithmic}[1]
		    \State Randomly sample number of steps to use waypoints generator $N^\text{waypoints-step}$ from $\{0, ... 15\}$
		    \For {$t = 0$ to $N^\text{waypoints-step} - 1$}
		        \State Generate waypoints using the waypoints generator as: $\bm{w}_t = G(I^\text{sim}_t, \bm{s}^\text{robot}_t)$
		        \State Reset robot state to the closest waypoints $\bm{w}_t$
		    \EndFor
			\State Start an episode with an initial state of $\bm{s}^\text{robot}_\text{waypoints-step}$
		\end{algorithmic}
	\end{algorithm}

\end{document}